\documentclass[twoside,11pt]{article}
\usepackage[abbrvbib,preprint]{jmlr2e}
\usepackage{amsmath}
\usepackage{bbm}
\usepackage{booktabs}
\usepackage{tabularx}
\usepackage{tikz}
\usetikzlibrary{positioning,arrows.meta,calc}
\usepackage{microtype}
\usepackage{lastpage}
\usepackage[T1]{fontenc}
\usepackage[utf8]{inputenc}
\jmlrheading{1}{2026}{1--\pageref{LastPage}}{8/26}{}{}{Michał Tomaszewski}
\ShortHeadings{Self-Interventional Learning}{Tomaszewski}
\firstpageno{1}
\begin{document}
\title{Can Neural Networks Learn by Experimenting on Themselves?\\
Self-Interventional Learning from Functional Consequences to Predictive Self-Knowledge}
\author{\name Michał Tomaszewski \email m.tomaszewski@po.edu.pl \\
\addr Department of Artificial Intelligence\\
Faculty of Computer Science\\
Opole University of Technology\\
Prószkowska 76, 45-758 Opole, Poland}
\maketitle
\begin{abstract}
Machine-learning systems usually model external data, while their internal functional organization is analyzed by external observers. This work introduces \emph{Self-Interventional Learning} (SIL), in which a neural system perturbs its own functional structure, observes consequences, learns a predictive self-model, generalizes to unexecuted interventions, and uses predictions to guide later structural action. In a construction-known synthetic system, SIL recovered critical structure, redundancy, and replaceability, while synergy was not reliably recovered. Across 30 fresh confirmatory seeds, increasing the pairwise intervention budget from 4 to 56 reduced held-out prediction error from 0.0335 to 0.0148 and increased Spearman correlation from 0.629 to 0.883. In a matched ablation, preserving the correct intervention--consequence mapping reduced prospective prediction error by 81.3\%, while using the same learned self-model for action reduced normalized regret by 31.7\% relative to ignoring it. However, model-guided action did not significantly outperform a direct empirical-memory policy, and powered CIFAR-10/ResNet validation showed no robustness advantage over equal-budget direct repair search. These results support SIL as an intervention-driven framework for learning predictive knowledge about a network's own functional organization, while showing that the self-model remains incomplete and is not universally superior to simpler direct strategies.
\end{abstract}
\section{Introduction}

Machine learning is predominantly outward-facing. Supervised systems learn mappings from observations to targets, self-supervised systems extract regularities from the data-generating process, reinforcement-learning agents learn value-relevant structure in an environment, and model-based agents learn predictive models of external dynamics. By contrast, the functional organization of the learner itself is usually treated as an object of analysis rather than an object of learning. Neural components are ablated, patched, ranked, visualized, or assembled into circuits by researchers and interpretability algorithms, but the resulting intervention evidence is rarely treated as experience from which the system acquires a predictive model of its own functional organization.

This separation is increasingly important because causal analysis of neural networks has made internal intervention a standard scientific instrument. Mechanistic interpretability uses ablation, activation patching, causal tracing, interchange intervention, path patching, and related methods to identify components that contribute causally to model behavior \cite{geiger2023causal,conmy2023automated,heimersheim2024patching}. These methods can expose structure that correlational probes cannot establish, and semi-synthetic benchmarks with known circuits have been developed to evaluate whether interpretability techniques recover ground truth \cite{gupta2024interpbench}. At the same time, recent analyses emphasize that singleton component importance can be insufficient: interaction effects may hide redundant, backup, or context-dependent mechanisms, and pairwise or higher-order intervention structure can be necessary to recover the operative computation \cite{vaidyanathan2026interactions}. In the standard interpretability setting, however, the experimenter learns from these interventions; the target network does not.

A distinct research tradition has studied self-modeling in embodied agents. Robots can learn predictive models of their own morphology or dynamics from sensorimotor interaction and then use those models for adaptation, damage recovery, planning, or behavior synthesis \cite{bongard2006resilient,kwiatkowski2019taskagnostic,kwiatkowski2022origins}. More recent work has also examined neural systems trained to predict their own internal states as an auxiliary objective, showing that self-modeling can alter the network itself by reducing complexity and increasing regularity \cite{premakumar2024unexpected}. These results establish that predictive models of the self can be useful. They do not, however, address the question considered here: can a neural learner experimentally intervene on its own computational substrate, learn the functional consequences of those interventions, generalize that knowledge to interventions it has not executed, and use the acquired self-knowledge to guide later structural decisions?

This work introduces \emph{Self-Interventional Learning} (SIL) as a framework for that problem. The central mechanism is
\[
\begin{aligned}
\text{self-intervention} &\rightarrow \text{observed functional consequence} \rightarrow \text{learned predictive self-knowledge}\\
&\rightarrow \text{generalization} \rightarrow \text{knowledge-guided action}.
\end{aligned}
\]
The object being modeled is neither an external environment nor only the current hidden state of the network. It is the causal-functional response of the learner's own structural components to controlled intervention. An intervention can remove, suppress, replace, or jointly manipulate components; the observed functional consequence becomes a training signal for a predictive self-model. The resulting model is then evaluated prospectively on interventions that were not executed during self-discovery. This framing makes self-knowledge empirically testable: it must predict counterfactual consequences, improve with additional self-interventional experience, and affect action when coupled to a decision policy.

The framework emerged from a structural-adaptation program referred to as the \emph{Kintsugi} line of experiments. Persistent structural scars were found to retain information from earlier perturbations, and causal/self-interventional models learned to predict the consequences of structural changes. This developed into Conditional Prospective Protection Value (CPPV), where candidate-conditioned repair value was predicted from self-interventional evidence. On Fashion-MNIST with an ElasticCNN, learned CPPV independently replicated an advantage over equal-budget direct repair evaluation and selected many repairs that had not been directly micro-trialed. A powered external replication on CIFAR-10 with a ResNet architecture produced a different and important result: protection prediction and extrapolation transferred, but the learned-vs-direct performance advantage was approximately zero. This boundary is retained rather than treated as a result to be tuned away. SIL is therefore not presented as a universally superior robustness method.

The final experimental program was designed to separate the broader learning claim from robustness performance. Experiment A constructs a neural system whose functional organization is known by design but hidden from the learner. Critical components, redundant pairs, replaceable components, synergistic/complementary pairs, and peripheral components can therefore be evaluated against ground truth. SIL recovered criticality, strongly recovered redundancy, and identified replaceability. Pairwise self-interventional experience reduced held-out intervention-prediction error relative to a singleton self-model. At the same time, synergy recovery failed its confirmatory hypothesis, demonstrating that the acquired self-model was informative but structurally incomplete. Importantly, critical-unit ranking alone did not distinguish SIL from classical importance methods: weight magnitude, gradient/Taylor importance, and direct singleton ablation could also identify critical units. The additional evidence lies in relational structure and prospective prediction rather than in scalar importance ranking.

Experiment B asks whether self-interventional knowledge behaves like learned knowledge rather than a fixed diagnostic. Across 30 fresh confirmatory seeds, the pairwise intervention budget was increased through a nested sequence while the singleton evidence and held-out evaluation set were held fixed. Held-out mean absolute error decreased systematically, rank correlation increased, and calibration improved. From 4 to 56 pairwise interventions, mean held-out error fell from 0.0335 to 0.0148, while Spearman correlation rose from 0.629 to 0.883. The preregistered slope hypotheses passed multiplicity correction, with the expected direction in 28/30 and 29/30 seeds, respectively. This supports an intervention-budget learning curve: more self-interventional experience produced better predictive self-knowledge. A heuristic informed-acquisition strategy did not improve sample efficiency over random exploration, and challenge-specific improvement did not survive multiplicity correction; these negative results delimit rather than invalidate the learning claim.

Experiment C tests whether the learned self-knowledge is causally relevant to action. A predictor-only negative control preserved the quantity and marginal distribution of pairwise evidence while permuting the mapping between intervention identity and consequence. The correctly paired self-model achieved a held-out mean absolute error of 0.0138 compared with 0.0740 for the permuted control, supporting the claim that predictive self-knowledge depends on the semantics of the intervention--consequence relationship rather than only on the target distribution. The same fitted self-model was then used in two action conditions: one ignored its predictions and selected actions randomly, while the other used the predictions to select a structural reconfiguration. Model-guided action reduced normalized regret from 0.3826 to 0.2614. Because these conditions share the same intervention history and fitted model, the contrast isolates the link from knowledge to action. Yet Full SIL did not significantly outperform a simpler direct empirical-memory policy. The advantage of learned relational knowledge was concentrated in interaction-sensitive redundancy decisions, while direct singleton-additive memory remained highly effective on matched-general cases and synergy remained difficult.

The resulting contribution is therefore narrower than a claim of universal algorithmic superiority but broader than a new post-hoc explanation method. To our knowledge, prior work has separately demonstrated embodied self-modeling, neural self-prediction, causal intervention for mechanistic analysis, and active experimental design for discovering external causal systems \cite{tigas2022interventions,ji2025survey}. SIL combines a different set of requirements: the learner's own functional substrate is the intervention target; intervention consequences become experience for a predictive self-model; self-knowledge is evaluated on genuinely unexecuted interventions; its quality is tested as a function of intervention experience; and its downstream role is isolated by causal action ablation. The central claim of this paper is consequently that self-intervention can serve as a source of predictive self-knowledge in neural systems, and that such knowledge can be learned, generalized, and used---without implying that it is complete or universally preferable to simpler direct strategies.

\paragraph{Contributions.}
The main contributions are:
\begin{enumerate}
    \item \textbf{Self-Interventional Learning framework.} A learning formulation in which controlled interventions on a system's own functional structure generate experience for a predictive self-model, linking self-intervention, consequence learning, held-out generalization, and structural action.
    \item \textbf{Ground-truth structural test.} A synthetic construction-known protocol that distinguishes scalar importance from redundancy, replaceability, interaction recovery, structural-role inference, and prediction of genuinely unexecuted interventions. The results show strong recovery of redundancy and replaceability while retaining a confirmatory failure for synergy.
    \item \textbf{Confirmatory self-knowledge learning curve.} A nested intervention-budget experiment showing that increasing relational self-interventional experience systematically reduces prediction error, improves ranking quality, and improves calibration across fresh seeds.
    \item \textbf{Causal intervention--knowledge--action ablation.} A negative-control permutation and matched action ablation showing that preserving the correct intervention--consequence mapping substantially improves prospective self-prediction and that using the same learned self-model for action reduces decision regret relative to ignoring it.
    \item \textbf{Applied instantiation with explicit external boundary.} The Kintsugi/CPPV experiments connect predictive self-knowledge to structural adaptation, while independent Fashion-MNIST replication and powered CIFAR-10/ResNet validation establish both an applied success case and a domain-dependent limit: predictive knowledge transfers more reliably than superiority over equal-budget direct search.
\end{enumerate}


\section{Related Work}
\label{sec:related_work}

Self-Interventional Learning (SIL) is positioned at the intersection of several research traditions: self-modeling systems, interpretable-by-design networks, mechanistic and intervention-based analysis, structural importance and topology adaptation, causal abstraction, and active causal discovery. Each of these traditions uses some form of internal modeling or intervention, but they differ in what is being modeled, who selects the intervention, what supervision is available, and whether the resulting knowledge is used by the same learner that generated it. SIL is defined by the conjunction of these properties: the intervention target is the learner's own functional substrate; intervention consequences are observed by the learner itself; a predictive self-model is learned from those consequences; generalization is evaluated on interventions not executed during self-discovery; and the learned self-knowledge can subsequently guide structural action.

\subsection{Self-modeling systems}

Self-modeling has a substantial history in adaptive robotics. Early work on continuous self-modeling demonstrated that a robot could infer a model of its own morphology from actuation--sensation experience and use that model to recover from structural change \cite{bongard2006resilient}. Later work developed task-agnostic self-modeling and visual self-modeling of robot morphology, showing that embodied agents can acquire models of their bodies without an explicit hand-coded kinematic description \cite{kwiatkowski2019taskagnostic,chen2022visual,ledezma2023selfdiscovery}. These systems establish an important precedent: an artificial agent can learn a representation of itself that is useful for prediction, planning, or adaptation.

The present work addresses a different object of self-modeling. SIL does not primarily estimate the external morphology or physical dynamics of an embodied agent. Instead, the learner treats its own internal functional organization as an experimental object. The relevant unknowns are relations such as criticality, redundancy, substitutability, interaction, and prospective functional consequence under structural intervention. Moreover, the central evidence is not merely that a self-model can exist, but that predictive self-knowledge improves systematically as self-interventional experience increases and that the same learned model can be causally coupled to subsequent structural action.

A complementary neural self-modeling line trains artificial networks to predict their own internal states as an auxiliary objective. Premakumar et al. showed that such self-prediction can alter network organization and induce self-regularizing effects \cite{premakumar2024unexpected}. This establishes that prediction of one's own internal state can itself be a meaningful learning objective. SIL differs in both supervision and target: it learns from experimentally induced functional consequences rather than from prediction of naturally occurring internal states, and it models counterfactual structural outcomes rather than reconstructing or forecasting latent activity. Thus, SIL is closer to intervention-driven system identification of the learner's own functional substrate than to auxiliary state prediction.

\subsection{Self-explaining and interpretable-by-design networks}

Self-explaining neural networks and related interpretable-by-design architectures seek to make model decisions more explicit, faithful, and stable by constraining the form of the predictor or by introducing semantically meaningful intermediate variables \cite{alvarezmelis2018senn}. Concept-based approaches provide complementary mechanisms: TCAV quantifies sensitivity to user-defined semantic concepts, whereas concept bottleneck models route predictions through human-specified concepts that can themselves be intervened on at test time \cite{kim2018tcav,koh2020cbm}. These approaches reduce the opacity of the mapping from input to prediction, but they do not generally require a learner to experimentally perturb its own structural components, infer the functional consequences of those perturbations, or construct a predictive model of its own organization. SIL therefore uses \emph{self-knowledge} in a different sense: the object being learned is not a human-facing explanation of an input--output decision, but a predictive model of how the learner's own function changes under intervention.

This distinction is important because a system can be interpretable without possessing a learned model of its own intervention response, and conversely a useful predictive self-model need not be directly human-semantic. The present experiments therefore evaluate predictive accuracy, calibration, structural-relation recovery, and decision regret rather than explanation plausibility or concept alignment.

\subsection{Mechanistic interpretability and intervention-based analysis}

Mechanistic interpretability spans both observational localization and explicit intervention. Network Dissection, for example, quantifies alignment between hidden units and semantic visual concepts, while sanity-check work has shown that explanation methods must be tested for sensitivity to the learned model rather than judged from visual plausibility alone \cite{bau2017networkdissection,adebayo2018sanity}. Intervention-based approaches go further: activation patching, interchange interventions, causal tracing, causal mediation, and automated circuit discovery replace or perturb internal states and measure changes in model behavior \cite{geiger2021causal,conmy2023automated,heimersheim2024patching,geiger2023causal,vig2020causalmediation,meng2022rome}. Together, these methods provide a strong conceptual precedent for treating neural components as experimental targets rather than relying only on observational attribution.

The key difference is the locus of agency and the purpose of the intervention. In conventional mechanistic interpretability, interventions are selected and interpreted by an external analyst to explain a trained system. SIL instead makes intervention evidence part of the learner's own learning process. The system observes the consequences of interventions on itself, fits a predictive self-model from those observations, and is evaluated on whether that model generalizes to unexecuted interventions and can guide subsequent action. The distinction is therefore not between causal and non-causal analysis, but between \emph{external causal interrogation} and \emph{internalized intervention-driven learning}.

Recent work also sharpens why scalar component importance is not sufficient for the present objective. Vaidyanathan et al. showed that activation-patching estimands contain interaction effects and that components whose importance depends on other components can be missed by greedy single-component ranking; pairwise and higher-order interactions can therefore be structurally consequential \cite{vaidyanathan2026interactions}. This observation is closely aligned with the motivation for the synthetic ground-truth experiment in this work, where singleton importance is deliberately separated from redundancy and complementarity. Importantly, this connection does not imply that SIL resolves all interaction structure: the confirmatory experiments recover redundancy strongly but do not establish reliable recovery of synergistic/complementary structure.

\subsection{Structural importance, pruning, and topology adaptation}

A second relevant tradition assigns importance to parameters or units in order to remove, preserve, or restructure parts of a network. Classical second-order pruning methods such as Optimal Brain Damage and Optimal Brain Surgeon estimate the loss increase associated with deleting parameters, while later magnitude- and gradient-based methods scale this logic to modern neural networks \cite{lecun1989obd,hassibi1993obs,han2015connections,molchanov2017pruning,sanh2020movement}. Dynamic sparse training further shows that network connectivity itself can be treated as an adaptive object during optimization rather than as a permanently fixed scaffold \cite{evci2020rigl}.

These methods are important baselines and conceptual neighbors because they operationalize structural saliency and structural change. Their objective, however, is usually compression, efficient optimization, or direct task performance. SIL asks a different question: whether outcomes of controlled structural interventions can become a training set for a predictive model of the network's own intervention response. This distinction is why Experiment A does not treat success on critical-unit ranking as sufficient evidence for SIL. Classical magnitude, gradient/Taylor, and direct-ablation scores can identify critical components; the additional target is relational structure and prediction of unexecuted interventions.

\subsection{Causal abstraction and interchange intervention training}

Causal abstraction provides a formal framework for asking whether a high-level causal model faithfully describes the internal computation of a neural network. Interchange interventions can be used to test whether aligned neural representations have the counterfactual properties required by the proposed abstraction \cite{geiger2021causal,geiger2023causal}. Interchange Intervention Training (IIT) goes further by training a neural model to realize a target high-level causal structure under aligned counterfactual interventions \cite{geiger2022iit}.

SIL differs from this framework in a crucial source-of-structure assumption. In causal-abstraction analysis and IIT, an interpretable high-level causal model or alignment hypothesis is supplied by the researcher and then tested or imposed. In the central SIL experiments, the functional organization is hidden from the learner. The learner is not given labels such as ``critical'', ``redundant'', ``replaceable'', or ``synergistic'', nor a symbolic causal model to realize. Instead, it must infer predictive regularities from intervention identities and observed functional consequences. The synthetic benchmark uses construction-known structure only as an evaluator-side ground truth, creating a firewall between what is known by the experimenter and what is available to the learner.

\subsection{Causal representation learning}

Causal representation learning asks how high-level causal variables and mechanisms can be identified from lower-level observations, often using structural assumptions or interventional data \cite{scholkopf2021causalrepresentation}. SIL shares the emphasis on intervention-derived structure but changes the target of inference: the object is not a causal representation of an external data-generating world, but the intervention response of the learner's own functional substrate. The present work therefore does not claim to discover a complete structural causal model of the network; it learns a predictive intervention--consequence map whose relational adequacy can be tested against construction-known structure.

\subsection{Active causal discovery and experimental design}

Active causal discovery studies how to select interventions that efficiently reduce uncertainty about an unknown structural causal model. Interventional Markov-equivalence theory formalizes how experiments refine causal identifiability, while experiment-selection methods ask which interventions should be performed under finite budgets \cite{hauser2012interventional,he2008activecausal,hyttinen2013experiment}. Later work develops adaptive strategies for restricted graph families, differentiable multi-target Bayesian experimental design, and amortized intervention policies learned across causal systems \cite{greenewald2019causaltrees,tigas2022interventions,tigas2023diffcbed,annadani2024caasl}. This research shares with SIL the premise that interventions can be treated as informative experiments rather than merely perturbations.

The difference is again the object of discovery. Active causal discovery conventionally assumes a learner or experimenter that acts on an external system of variables. SIL treats the learner's own functional substrate as that system. In addition, the primary intervention-budget experiment in this work intentionally uses nested random acquisition, rather than optimized intervention selection, so that the causal variable under study is the \emph{amount of self-interventional experience}. A secondary informed acquisition heuristic did not improve sample efficiency, reinforcing the decision not to equate SIL with active experimental-design optimization.

\subsection{Position of Self-Interventional Learning}

Taken together, prior work establishes that artificial systems can model themselves, that neural structure can be ranked or adaptively modified, that internal mechanisms can be probed through intervention, that causal structure can be tested through counterfactual alignment, and that interventions can be selected to learn unknown causal systems. SIL combines a different set of commitments. It treats a learner's own functional structure as an intervention domain; converts observed consequences into a predictive self-model; evaluates that model on genuinely unexecuted interventions; measures whether self-knowledge improves with additional interventional experience; and tests whether the acquired self-model changes action when coupled to a decision policy.

To our knowledge, the distinctive contribution is therefore not the first use of self-models, interventions, causal reasoning, or internal prediction in isolation. It is the formulation and confirmatory evaluation of an intervention-driven learning process in which a neural system experimentally interrogates its own functional substrate and converts those experiments into predictive self-knowledge that can generalize and guide action. The empirical results also delimit this claim: relational redundancy is recoverable, synergistic structure remains difficult, and learned model-guided action is not universally superior to a direct empirical intervention policy.


\section{Self-Interventional Learning}
\label{sec:sil_framework}

\subsection{Problem setting}

Consider a learner with parameters or functional components collected in a substrate
\[
\mathcal{S}=\{s_1,\ldots,s_m\},
\]
and a task-dependent functional response
\[
f_{\theta}:\mathcal{X}\rightarrow\mathcal{Y}.
\]
An intervention is an operator
\[
I\in\mathcal{I}(\mathcal{S})
\]
that modifies the learner's own functional substrate without changing the external input distribution. Examples include lesioning a component, jointly lesioning several components, substituting one component for another, or applying another predeclared structural transformation.

For an input distribution \(P_X\), let the functional consequence of intervention \(I\) be
\[
c(I)=D\!\left(f_{\theta},f_{\theta}^{I};P_X\right),
\]
where \(f_{\theta}^{I}\) denotes the intervened system and \(D\) is a predefined functional discrepancy. The specific discrepancy may be task-dependent; in the synthetic experiments it is defined on unlabeled probe inputs so that the self-discovery process does not require task labels.

A self-interventional experience is then a tuple
\[
e_t=(I_t,c(I_t)),
\]
and the experience set after \(T\) interventions is
\[
\mathcal{E}_T=\{e_1,\ldots,e_T\}.
\]

\subsection{Predictive self-knowledge}

SIL introduces a predictive self-model
\[
g_{\phi}:\mathcal{I}(\mathcal{S})\rightarrow\mathbb{R}
\]
trained from \(\mathcal{E}_T\) to estimate the consequence of a candidate intervention:
\[
\hat c(I)=g_{\phi}(I\mid\mathcal{E}_T).
\]
The self-model is not defined by its architecture but by its epistemic role: it maps descriptions of interventions on the learner's own substrate to predicted functional consequences. The learned object is therefore a predictive model of intervention response rather than a human-semantic explanation label.

The central generalization requirement is evaluated on a held-out set
\[
\mathcal{H}\subset\mathcal{I}(\mathcal{S}),\qquad
\mathcal{H}\cap\mathcal{I}_{\mathrm{executed}}=\emptyset,
\]
whose true consequences are not executed before model freeze. Predictive self-knowledge is operationally supported when
\[
\hat c(I),\; I\in\mathcal{H},
\]
tracks the subsequently revealed \(c(I)\) according to preregistered error, ranking, and calibration metrics.

\subsection{Relational functional structure}

Singleton importance is insufficient when functional consequences depend on interactions between components. For components \(s_i\) and \(s_j\), define
\[
J_{ij}=c(I_{ij})-c(I_i)-c(I_j),
\]
where \(I_i\) and \(I_j\) are singleton interventions and \(I_{ij}\) is their joint intervention. Under the synthetic construction used here, large positive \(J_{ij}\) corresponds to redundancy-like dependence: neither singleton intervention is highly consequential, but the joint intervention is. Negative \(J_{ij}\) corresponds to complementary/synergistic structure under the adopted consequence definition. Replaceability is evaluated separately through directed substitution interventions.

This distinction separates three epistemic levels:
\begin{enumerate}
    \item \textbf{scalar importance}: estimating the consequence of individual components;
    \item \textbf{relational self-knowledge}: estimating how consequences change under joint or substitutive interventions;
    \item \textbf{predictive self-modeling}: generalizing these estimates to interventions not executed during self-discovery.
\end{enumerate}

\subsection{Learning from self-interventional experience}

A framework qualifies as a learning process only if predictive self-knowledge changes systematically with experience. Let \(B\) denote an intervention budget and let \(g_{\phi_B}\) be the self-model trained from a nested experience set \(\mathcal{E}_B\), with
\[
\mathcal{E}_{B_1}\subset\mathcal{E}_{B_2}\quad\text{for}\quad B_1<B_2.
\]
The learning-efficiency hypothesis is that predictive quality improves as \(B\) increases. For a prediction error \(L_B\), this implies a negative within-seed trend,
\[
\frac{\partial L_B}{\partial \log B}<0,
\]
while for a rank-quality measure \(Q_B\),
\[
\frac{\partial Q_B}{\partial \log B}>0.
\]
These trends are evaluated across independent seeds rather than by treating intervention-level observations as independent replicates.

\subsection{Knowledge-guided action}

Predictive self-knowledge becomes behaviorally consequential when it controls a subsequent action. Given a candidate action set \(\mathcal{A}\subset\mathcal{I}(\mathcal{S})\), a model-guided policy selects
\[
a^{\mathrm{SIL}}=\arg\min_{a\in\mathcal{A}} g_{\phi}(a),
\]
when lower predicted consequence is preferable. The realized normalized regret is
\[
R(a)=\frac{c(a)-\min_{a'\in\mathcal{A}}c(a')}
{\max_{a'\in\mathcal{A}}c(a')-\min_{a'\in\mathcal{A}}c(a')+\varepsilon}.
\]
A causal knowledge-to-action test requires holding the learned self-model fixed while changing only whether its predictions are used by the policy. In the confirmatory ablation, the knowledge-disconnected and knowledge-guided conditions share the same intervention history, fitted model, candidate set, and predictions; they differ only in action selection.

\subsection{Operational definition of SIL}

A system is treated as exhibiting Self-Interventional Learning in the present framework when the following properties are jointly testable:
\begin{enumerate}
    \item \textbf{Self-intervention}: the system obtains training evidence by controlled interventions on its own functional substrate.
    \item \textbf{Consequence observation}: each executed intervention produces an observed functional consequence under a predefined measurement rule.
    \item \textbf{Predictive self-modeling}: intervention--consequence experience is converted into a model that predicts outcomes of candidate self-interventions.
    \item \textbf{Held-out generalization}: the self-model predicts interventions whose true consequences were not executed before model freeze.
    \item \textbf{Experience-dependent improvement}: predictive self-knowledge improves systematically as self-interventional experience increases.
    \item \textbf{Knowledge-guided action}: the learned self-model can be coupled to a policy so that predictions influence a subsequent structural decision.
\end{enumerate}

This definition is intentionally narrower than claims of self-awareness or autonomous introspection. SIL concerns experimentally acquired predictive knowledge about a system's own functional intervention response. It also does not require universal superiority over direct empirical strategies. A simple intervention memory may be optimal in approximately additive regions of the intervention space, whereas a learned relational self-model is expected to be most useful when generalization across unmeasured, interaction-sensitive interventions is required.

\subsection{Evidence map and scope of the present study}

The empirical program tests these requirements in three final mechanistic experiments. Experiment A asks whether known but learner-hidden functional structure can be recovered from self-interventional evidence. Experiment B tests whether predictive self-knowledge improves with a preregistered increase in relational intervention budget. Experiment C breaks the intervention--knowledge--action chain and tests whether preserving intervention semantics improves prediction and whether coupling the learned self-model to action reduces prospective decision regret.

The framework is additionally instantiated in the earlier Kintsugi/CPPV experiments, where self-interventional consequence models are used for structural repair selection. Those experiments demonstrate practical transfer and also establish an external boundary: learned repair selection is superior to equal-budget direct search in the Fashion-MNIST/ElasticCNN setting, while a powered CIFAR-10/ResNet replication shows no corresponding performance advantage despite preserved predictive protection signals. The framework claim is therefore about the existence and use of intervention-derived predictive self-knowledge, not universal robustness superiority.


\section{Experimental Program and Methods}
\label{sec:methods}

\subsection{Study logic}
\label{sec:methods_logic}

The empirical program was organized to distinguish an applied structural-adaptation effect from the broader claim that intervention on a learner's own functional substrate can generate predictive self-knowledge. The first lineage, termed Kintsugi/CPPV, provided an applied instantiation in image-classification networks. Three final mechanistic experiments then isolated progressively stronger requirements of Self-Interventional Learning (SIL): recovery of construction-known but learner-hidden functional structure (Experiment A), experience-dependent improvement of predictive self-knowledge (Experiment B), and causal use of the learned self-model for prospective action (Experiment C). The latter experiments were intentionally synthetic because their purpose was not benchmark performance but identifiability: the evaluator required exact knowledge of criticality, redundancy, complementarity, replaceability, and peripheral function while keeping those roles hidden from the learner.

Across all confirmatory studies, the seed was the independent replication unit. Pilot runs were used only for technical validation, protocol promotion, and correction of defects identified before confirmation; pilot outcomes were not pooled into confirmatory inference. Confirmatory seed sets were fresh unless the relevant frozen protocol explicitly defined a matched within-seed comparison. Primary endpoints, intervention budgets, holdout rules, decision gates, and multiplicity families were frozen before confirmatory outcomes were inspected. The resulting design therefore evaluates a sequence of increasingly specific claims rather than treating all experiments as interchangeable demonstrations of robustness.

\subsection{Common intervention and firewall principles}
\label{sec:common_firewall}

SIL separates ordinary task learning from self-interventional learning. Where a task model required supervised pretraining, labels were used only in that initial task-learning stage. Subsequent self-intervention, self-model fitting, structural selection, and repair-selection procedures operated on unlabeled buffers or probe inputs. Final task labels and evaluator-only structural roles were not available to the self-discovery process.

An intervention was treated as an experiment on the learner's own functional substrate. Its consequence was measured by a predeclared functional discrepancy rather than by a human-assigned structural label. In the synthetic experiments this discrepancy was entirely label-free; in the applied Kintsugi lineage, adaptation and repair evaluation likewise used label-free intervention evidence while final held-out lesion families remained unavailable to adaptation. The central firewall principle was stronger than ordinary train--test separation: for held-out structural interventions, the true consequence itself was not executed before the relevant model or decision policy was frozen. Thus, generalization was evaluated on interventions that were genuinely unperformed during self-discovery, not merely on outcomes withheld from a fitted regressor.

Where structural ground truth existed, role names, latent construction identifiers, true pair identities, and oracle interaction signs were evaluator-side metadata only. Acquisition policies could use only predeclared learner-visible descriptors such as component topology, activation statistics, singleton consequences, or deterministic randomization. Every final experiment stored an auditable intervention catalog and verified zero overlap between discovery interventions and forced challenge interventions before evaluation.

\subsection{Applied instantiation: Kintsugi structural adaptation}
\label{sec:kintsugi_methods}

The Kintsugi lineage used Fashion-MNIST with an ElasticCNN containing reserve structural capacity. The task model was first trained conventionally. The self-interventional stage then received an unlabeled SIL buffer. In the frozen confirmatory protocol, SIL-K performed 80 adaptation interventions after 16 warm-up episodes and explored four predeclared intervention families: singleton and pair lesions in the second convolutional layer, singleton lesions in the first fully connected layer, and sparse multi-unit lesions in that layer. A Bayesian Causal Self-Model (CSM) was fit to intervention--consequence observations. Persistent structural opportunities occurred at episode 20 and every ten episodes thereafter, yielding at most seven scars. Scar targets required at least one actual singleton observation, and candidate ranking used the posterior consequence of the canonical singleton intervention with a conservative lower-confidence adjustment. Persistent scars were implemented through function-preserving duplication followed by label-free consolidation.

The information budget of the strongest direct structural comparator was explicitly matched. Importance-based regrowth received exactly 80 measured probe interventions and the same seven structural opportunities as SIL-K. Additional controls included the frozen pretrained network, dropout training, random damage, self-distillation repair, and random regrowth. The final damage families, \texttt{block\_conv2} and \texttt{mixed\_cross\_layer}, were excluded from adaptation. They were evaluated at severities 0.30, 0.45, 0.60, and 0.75 with matched lesion randomization across methods. The primary endpoint was mean accuracy retention, defined as damaged accuracy divided by the matched clean accuracy, with absolute damaged accuracy retained as an important audit because retention can be affected by changes in the clean denominator.

\subsection{Conditional Prospective Protection Value}
\label{sec:cppv_methods}

The Kintsugi mechanism was extended from estimating structural vulnerability to estimating the prospective value of a candidate repair. Conditional Prospective Protection Value (CPPV) used reversible micro-trials at the seven repair checkpoints. Each conditional method received four reversible candidate-repair trials per checkpoint, for a maximum of 28 repair micro-trials. Candidate probing was layer-balanced and random with respect to the learned protection score, so the comparison between learned and direct selection tested generalization of repair value rather than an active-search advantage.

For a candidate repair, a predeclared conditional stress panel always contained the candidate source and used only adaptation-side stress generators distinct from the final held-out lesion families. Tail-aware risk was
\[
\operatorname{Risk}=0.25\,\operatorname{mean}+0.75\,\operatorname{CVaR}_{0.75},
\]
and normalized protection gain was
\[
G=\frac{\operatorname{Risk}_{\mathrm{pre}}-\operatorname{Risk}_{\mathrm{post}}}
{\operatorname{Risk}_{\mathrm{pre}}+\varepsilon}.
\]
Clean-function preservation was not subtracted from this target. Instead, it was enforced as a hard feasibility condition using a Jensen--Shannon discrepancy threshold of 0.002 on clean outputs. The learned policy selected the candidate with the largest lower-confidence protection estimate, \(\widehat G-0.25\widehat\sigma_G\), subject to feasibility; the equal-budget direct policy selected among the candidates actually measured by reversible micro-trials. The full method was compared with frozen, dropout-trained, importance-regrowth, SIL-K singleton-only, direct CPPV, and a learned CPPV ablation without CSM-derived features.

The independently replicated Fashion-MNIST protocol froze the same algorithmic constants and used 35 new seeds (150 to 388 with spacing seven). Primary endpoints were mean damaged accuracy and mean retention. Learned CPPV was compared with equal-budget Direct CPPV and with SIL-K singleton-only in a four-test Holm family. Phase data from the earlier confirmatory study were treated as background evidence only and were not pooled into the replication statistics.

\subsection{External CIFAR-10/ResNet validation}
\label{sec:cifar_methods}

External validation replaced Fashion-MNIST/ElasticCNN with CIFAR-10 and an ElasticCifarResNet20 while retaining the CPPV logic, held-out evaluation concept, and matched learned-versus-direct comparison. The first transfer attempt exposed an architecture-specific consolidation failure: global consolidation updates and BatchNorm running-statistic changes made reversible repairs violate the frozen clean-function safeguard. This run was retained as a transfer diagnostic rather than pooled into later inference. Before the architecture-safe external study, consolidation was corrected by freezing BatchNorm running statistics and restricting updates to scar-local parameters. The correction was frozen before new external seeds were evaluated and did not use final held-out lesion outcomes for tuning.

The final powered external replication used 82 fresh seeds (611, 618, \ldots, 1178) and primary inference only from those seeds. Learned and Direct CPPV received the same micro-trial budget and structural opportunities. The same two primary endpoints---mean damaged accuracy and mean retention---were tested under the frozen multiplicity rule. Architecture-safe feasibility, clean-function drift, protection-model quality, selection of candidates not directly micro-trialed, and damage burden were recorded as prespecified diagnostics. The purpose of this external phase was to distinguish transfer of predictive self-interventional knowledge from transfer of any performance advantage over equal-budget direct repair search.

\subsection{Synthetic functional system shared by Experiments A--C}
\label{sec:synthetic_system}

Experiments A--C used a construction-known modular neural system whose organization was hidden from the learner. For every seed,
\[
x\sim\mathcal{N}(0,I_{12}),
\]
and a fixed modular circuit mapped \(x\) to a binary logit. Task labels were generated only to verify that the intact system defined a nontrivial classification problem,
\[
y=\mathbbm{1}[\ell(x)+\epsilon>0],\qquad \epsilon\sim\mathcal{N}(0,0.35^2),
\]
but these labels were never available to SIL. The 24 functional components were anonymized by a seed-specific permutation to public identifiers \(U00,\ldots,U23\).

The construction contained three critical components, three redundant pairs, three complementary/synergistic pairs, two replaceable target--donor pairs, three peripheral components, and two neutral components. Critical components made large unique contributions. A redundant module remained functional if either member survived, so singleton lesion consequences were small whereas lesioning both removed the module. A complementary module contributed only while both members remained active, producing a negative interaction under the adopted consequence convention. A replaceable target was active while its donor was a standby functional duplicate; target lesion was consequential and a directed donor-to-target substitution restored the corresponding function. Peripheral components were deliberately weak and neutral components had no outgoing functional contribution.

For an intervention \(I\), functional consequence was measured on an unlabeled probe set as
\[
C(I)=\frac{\mathbb{E}_x[(p_0(x)-p_I(x))^2]}{\operatorname{Var}_x[p_0(x)]},
\]
where \(p_0\) and \(p_I\) are intact and intervened output probabilities. For a pair lesion,
\[
J(i,j)=C(\{i,j\})-C(\{i\})-C(\{j\}).
\]
Positive \(J\) is the protocol signature of redundancy and negative \(J\) the signature of complementarity/synergy. Before any learner ran, an oracle-only construction audit verified that each seed satisfied the predeclared inequalities defining the five role families. A seed failing these inequalities was a technical protocol invalidation and could not be silently replaced.

\subsection{Experiment A: synthetic structural ground truth}
\label{sec:method_A}

Experiment A asked whether a learner could recover known functional organization from controlled interventions on itself. Discovery allowed singleton lesions, pair lesions, and directed substitution interventions. All 24 singleton lesions were measured. Pairwise discovery used a fixed structural-blind budget selected from learner-visible evidence: 40\% of the budget emphasized topology overlap, 25\% absolute activation correlation, 20\% the sum of singleton consequences, and 15\% seeded random exploration. Directed substitutions were selected from generic activation and topology similarity rather than rescue outcome.

The firewall forced one true redundant pair, one true synergy pair, and one true replaceable relation into a challenge holdout for every seed. In addition, 25\% of remaining pair lesions and 50\% of substitution candidates were held out by a seed-stable hash. Their consequences were executed only after the SIL model was frozen. Seven methods were compared: random ranking, weight magnitude, activation magnitude, gradient/Taylor importance \cite{molchanov2017pruning}, direct singleton ablation, a SIL singleton model, and SIL with singleton plus selected pairwise evidence. Scalar baselines received no pairwise outcomes. Direct singleton used measured singleton consequence as its unit score and an additive extension to pair prediction; SIL singleton fit a regularized predictor from singleton intervention descriptors; SIL pairwise received the selected relational evidence and predicted unseen interventions from generic component and pair descriptors.

The five-seed pilot (1101--1105) used 8,000 probe inputs, a 40,000-sample construction audit, 40 pair lesions, and six substitution interventions. Its role was technical only. Confirmatory inference used 20 fresh seeds (2101--2120), 12,000 probe inputs, a 60,000-sample construction audit, 56 pair lesions, and eight substitutions, for 88 discovery interventions per seed. Four primary hypotheses were frozen: critical-unit AUROC above 0.5; redundancy average precision above the direct singleton additive baseline; synergy average precision above the same baseline; and lower held-out intervention MAE for SIL pairwise than SIL singleton. These four tests formed one Holm family at family-wise \(\alpha=0.05\).

\subsection{Experiment B: intervention-budget learning curve}
\label{sec:method_B}

Experiment B isolated the amount of relational self-interventional experience. The 24 singleton interventions were fixed for every learner, while pairwise experience increased through the predeclared sequence
\[
B\in\{4,8,14,28,56\},
\]
with \(B=0\) retained as a descriptive singleton-only anchor. No substitution interventions were used. The primary acquisition policy, \texttt{random\_nested}, generated one deterministic uniform-without-replacement ordering of eligible non-held-out pairs per seed; each budget was a strict prefix of that ordering. Consequently,
\[
\mathcal{E}_{4}\subset\mathcal{E}_{8}\subset\mathcal{E}_{14}\subset\mathcal{E}_{28}\subset\mathcal{E}_{56},
\]
so differences across budgets reflected added experience rather than different samples at each point. A secondary \texttt{informed\_nested} policy ranked candidates from learner-visible pre-pair information only, using topology cosine similarity, absolute activation correlation, singleton-consequence sum, and deterministic randomization with frozen weights 0.40/0.25/0.20/0.15. Neither policy adapted its ordering to observed pair outcomes.

The relational predictor was a \texttt{HistGradientBoostingRegressor} with maximum depth 4, learning rate 0.06, 180 boosting iterations, \(L_2\) regularization 1.0, and \texttt{min\_samples\_leaf}=2. The latter value was frozen before any Experiment B outcome was observed to avoid a purely algorithmic leaf-size degeneracy at the smallest sample sizes. Every budget model was trained from scratch; warm starts were prohibited. The holdout set was identical across budgets and acquisition policies within a seed and contained a seed-hash subset of pairs plus one forced redundant and one forced synergy challenge. No held-out consequence was executed until all budget-specific models for that seed had written frozen prediction artifacts.

The primary metrics were held-out pair MAE and held-out pair Spearman correlation. Within each seed, ordinary least-squares slopes were computed over \(\log_2 B\) for the five primary budgets. The first primary hypothesis required a negative MAE slope. The second required a positive slope of Fisher-\(z\)-transformed Spearman correlation. Confirmatory inference used 30 fresh seeds (4101--4130), 12,000 probe inputs, and a 60,000-sample construction audit. One-sided Wilcoxon signed-rank tests of per-seed slopes against zero formed a two-test Holm family. Calibration, challenge error, monotonicity, area under the learning curve, B50/B90, structural-recovery metrics, and informed-versus-random acquisition were predeclared secondary analyses and could not rescue a failed primary learning-curve result.

\subsection{Experiment C: intervention--knowledge--action causal ablation}
\label{sec:method_C}

Experiment C tested whether intervention-derived predictive self-knowledge was causally relevant to prospective structural action. It reused the same synthetic generator and the Experiment B relational learner. Self-interventional conditions received 24 singleton and 56 random-nested pair interventions, for 80 discovery interventions per seed. The downstream task consisted of nine two-component reconfiguration episodes. Each episode presented six candidate pairs and required choosing exactly one pair to deactivate temporarily while minimizing functional consequence. Three episodes were redundancy-sensitive, three synergy-sensitive, and three matched-general. Candidate sets were constructed from intervention identities and singleton-additive scores only; true held-out pair consequences were unavailable at decision time.

Four action conditions broke different links of the proposed SIL chain. C0 had no self-interventional knowledge and selected a precommitted random candidate. C1 received the self-intervention history but no learned relational predictor and selected the candidate with minimum direct singleton-additive score \(C(i)+C(j)\). C2 had the same intervention history and the same fitted self-model as the full method but ignored its predictions and used exactly the same precommitted random candidate as C0. C3, Full SIL, used the frozen self-model and selected the candidate with minimum predicted consequence. Thus the C3--C2 contrast held the intervention history, fitted model, candidate set, and predictions fixed and changed only whether the learned self-knowledge controlled action.

A predictor-only negative control tested the intervention-to-knowledge link. Singleton evidence was kept unchanged, while the 56 observed pair consequences were deterministically permuted among the pair rows before fitting an otherwise identical learner. This preserved sample count, model class, singleton evidence, and the marginal pair-target distribution while destroying the mapping from pair identity to observed consequence.

A strict commit barrier separated prediction from outcome revelation. Before any held-out candidate consequence could be executed, discovery had to be complete; the true and permuted predictors had to be fit and frozen; all held-out predictions had to be stored; all C0--C3 action choices had to be committed and hashed; and C0/C2 action identity had to be verified episode by episode. Only then were the unique held-out candidate interventions executed, and the oracle action was evaluated last. The primary action endpoint was seed-mean normalized regret,
\[
R(a)=\frac{C(a)-C(a_{\mathrm{best}})}{C(a_{\mathrm{worst}})-C(a_{\mathrm{best}})+10^{-8}}.
\]
Confirmatory inference used 30 fresh seeds (6101--6130), 12,000 probe inputs, and a 60,000-sample construction audit. The three primary hypotheses were: lower held-out candidate MAE for the true than the permuted self-model; lower regret for C3 than C2; and lower regret for C3 than C1. One-sided paired Wilcoxon tests at the seed level formed one three-test Holm family.

\subsection{Reproducibility and protocol provenance}
\label{sec:reproducibility_methods}

Experiments A--C were executed from protocol-locked repositories containing machine-readable protocol files, protocol hashes, construction validators, firewall validators, deterministic resume state, automated summaries, statistical audits, and lightweight result packagers. Protocol identifiers were \texttt{SIL-A-SYNTHETIC-GT-1.0.0}, \texttt{SIL-B-LEARNING-CURVE-1.0.0}, and \texttt{SIL-C-INTERVENTION-KNOWLEDGE-ACTION-1.0.0}. The corresponding protocol locks were retained with the result archives. Large model checkpoints were excluded from transfer packages by default; numerical outputs, configuration files, logs, frozen prediction artifacts, and statistical summaries were retained. Technical hotfixes discovered before confirmatory runs were limited to implementation defects or firewall enforcement and did not change frozen scientific hypotheses after outcomes were inspected.


\section{Statistical Discipline and Evidence Freezing}
\label{sec:stats_discipline}

\subsection{Independent replication unit and paired design}

The seed, not an intervention, lesion trial, candidate repair, or decision episode, was the independent inferential unit. Repeated interventions within a seed were used to construct seed-level predictive or decision metrics. Where methods shared a pretrained model, synthetic construction, held-out catalog, or action candidate set, inference used paired seed-level differences. This avoids pseudoreplication from treating multiple interventions generated by the same network instance as independent observations.

For paired comparisons, the analysis reported the mean and median difference, a nonparametric bootstrap 95\% confidence interval over seed-level paired differences, wins/losses/ties, and the paired standardized effect
\[
d_z=\frac{\overline{\Delta}}{s_{\Delta}}.
\]
Wilcoxon signed-rank tests were used for the frozen nonparametric primary comparisons. Directional one-sided tests were used only where the protocol specified an ordered hypothesis before confirmatory outcomes; otherwise comparisons remained two-sided as frozen in the corresponding lineage.

\subsection{Multiplicity control}

Multiplicity was handled within predeclared scientific families rather than by pooling unrelated tests. The Experiment A family contained four primary hypotheses (criticality, redundancy, synergy, and held-out prediction). Experiment B contained two primary learning-curve slope hypotheses. Experiment C contained three primary causal-ablation hypotheses. Each family used Holm correction with family-wise \(\alpha=0.05\). The Fashion-MNIST CPPV independent replication used its frozen four-test family spanning two endpoints and two controls. External CIFAR-10/ResNet inference retained its own frozen primary family and did not pool seeds or \(p\)-values from earlier Fashion-MNIST or diagnostic external runs.

Secondary endpoints and secondary Holm families were interpreted separately. They could refine mechanism or delimit scope but could not convert a failed primary gate into a positive confirmatory conclusion. In particular, severe-damage subfamilies, CSM-feature ablations, calibration analyses, challenge-specific metrics, informed acquisition, and structural-role decompositions were not allowed to rescue failed primary claims.

\subsection{Pilot--confirmatory separation}

Pilot stages served three purposes: software validation, verification that the intended signal was identifiable in principle, and discovery of technical defects before confirmation. Promotion decisions were based on frozen technical gates unless a protocol explicitly defined a directional pilot criterion before outcome inspection. Pilot observations were never combined with confirmatory seeds for final inference. When a genuine technical defect was discovered, the correction was documented before the confirmatory stage and the scientific endpoint family remained unchanged unless the corresponding protocol was explicitly superseded before confirmation.

This distinction was especially important in the Kintsugi lineage. An early pilot exposed confounded unit-level attribution and an unmatched information budget for importance-based regrowth. Both were corrected before the 20-seed confirmatory protocol: scars required direct singleton evidence and the strongest direct comparator was capped at the same 80 observed interventions as SIL-K. Similarly, the first CIFAR-10 transfer exposed an architecture-specific consolidation failure; that run was retained as a diagnostic and not pooled with the later architecture-safe external inference.

\subsection{Label, intervention, and decision firewalls}

Three firewall levels were used. First, after supervised task pretraining, self-interventional adaptation did not receive task labels. Second, final held-out damage families or evaluator-side structural roles were unavailable to the adaptation and self-modeling procedures. Third, in Experiments A--C, the true consequence of a held-out intervention was not executed before model freeze; Experiment C strengthened this further by requiring all action decisions to be committed and hashed before any candidate outcome was revealed.

The firewall therefore separates not only labels but also experimental knowledge. A learner could know that a candidate intervention existed and could compute predeclared descriptors for it, but it could not observe the outcome used for final evaluation. Challenge relations were forced outside discovery, and automated audits verified zero discovery--challenge overlap. This design makes prospective prediction of held-out interventions a genuine generalization test rather than retrospective interpolation over already executed experiments.

\subsection{Evidence freeze and stopping rule}

The empirical program used explicit no-rescue rules. Confirmatory seeds were not excluded because of unfavorable outcomes. Primary endpoints, thresholds, and multiplicity families were not changed after inspection. Earlier Fashion-MNIST and CIFAR-10 conclusions were frozen before Experiments A--C, including negative findings. The final three experiments were likewise frozen before execution, and a hard stop was imposed after Experiment C: no fourth benchmark, seed escalation, robustness dataset, or synergy-specific retuning was permitted unless a genuine technical error invalidated a frozen experiment.

Accordingly, the final article reports negative evidence as part of the claim boundary. Failure to recover synergy reliably in Experiment A, failure of informed acquisition to improve sample efficiency in Experiment B, failure of Full SIL to outperform direct empirical memory in Experiment C, lack of generic superiority over dropout, and the powered CIFAR-10 Learned-versus-Direct null are retained in the manuscript rather than treated as targets for post-hoc optimization.


\section{Results}
\label{sec:results}

\begin{table}[!htbp]
\centering
\small
\caption{Final experimental program. Pilot studies were used only for technical promotion; inferential conclusions below refer to fresh confirmatory seeds.}
\label{tab:program}
\begin{tabularx}{\textwidth}{p{24mm}Xp{25mm}p{25mm}p{30mm}}
\toprule
Block & Primary scientific question & Confirmatory unit & Primary family & Final decision \\
\midrule
Applied Kintsugi/CPPV & Does predictive self-interventional repair knowledge improve structural adaptation, and does that advantage transfer externally? & Fashion $N=35$ replication; CIFAR $N=82$ powered external & Frozen paired endpoint families & Fashion advantage replicated; CIFAR Learned$\approx$Direct performance boundary; predictive mechanism transfers \\
Experiment A & Can a learner recover construction-known but hidden functional structure and predict interventions it did not execute? & 20 fresh seeds & A-H1--A-H4; Holm FWER .05 & 3/4 primary PASS; synergy recovery FAIL \\
Experiment B & Does predictive self-knowledge improve systematically with increasing self-interventional experience? & 30 fresh seeds & B-H1--B-H2; Holm FWER .05 & 2/2 primary PASS \\
Experiment C & Are intervention--consequence semantics necessary for prediction, and does learned self-knowledge improve action when used? & 30 fresh seeds & C-H1--C-H3; Holm FWER .05 & 2/3 primary PASS; Full SIL $>$ direct memory not supported \\
\bottomrule
\end{tabularx}
\end{table}

\begin{table}[!htbp]
\centering
\scriptsize
\caption{Preregistered confirmatory hypotheses for Experiments A--C. Effect signs follow the stated contrast; negative values are favorable for MAE/regret hypotheses. All failures are retained in the main paper.}
\label{tab:primary}
\resizebox{\textwidth}{!}{%
\begin{tabular}{lllllll}
\toprule
ID & Contrast & $N$ & Mean effect & Bootstrap 95\% CI & Holm $p$ & Decision \\
\midrule
A-H1 & Critical AUROC $>0.50$ & 20 & $+0.5000$ & $[0.5000,0.5000]$ & $1.16\times10^{-5}$ & PASS \\
A-H2 & Redundancy AP: pairwise SIL $>$ Direct & 20 & $+0.53020$ & $[+0.35401,+0.70735]$ & $9.55\times10^{-5}$ & PASS \\
A-H3 & Synergy AP: pairwise SIL $>$ Direct & 20 & $+0.00558$ & $[+0.00040,+0.01267]$ & $0.1942$ & FAIL \\
A-H4 & Held-out MAE: pairwise SIL $<$ singleton SIL & 20 & $-0.01415$ & $[-0.01633,-0.01166]$ & $7.63\times10^{-6}$ & PASS \\
B-H1 & MAE slope vs. $\log_2(B)<0$ & 30 & $-0.004841$ & $[-0.006571,-0.003265]$ & $1.86\times10^{-8}$ & PASS \\
B-H2 & Fisher-$z$ Spearman slope vs. $\log_2(B)>0$ & 30 & $+0.151065$ & $[+0.113044,+0.190683]$ & $2.33\times10^{-8}$ & PASS \\
C-H1 & True-model MAE $<$ permuted-model MAE & 30 & $-0.06013$ & $[-0.06524,-0.05502]$ & $2.79\times10^{-9}$ & PASS \\
C-H2 & Full SIL regret $<$ same-knowledge/random-action regret & 30 & $-0.12116$ & $[-0.17212,-0.07035]$ & $2.32\times10^{-4}$ & PASS \\
C-H3 & Full SIL regret $<$ direct-memory regret & 30 & $-0.02448$ & $[-0.07539,+0.02980]$ & $0.2449$ & FAIL \\
\bottomrule
\end{tabular}%
}
\end{table}

\begin{table}[!htbp]
\centering
\small
\caption{Frozen evidence boundaries after completion of the experimental program.}
\label{tab:boundaries}
\begin{tabularx}{\textwidth}{p{48mm}p{30mm}X}
\toprule
Claim & Status & Evidence boundary \\
\midrule
Self-intervention consequences form predictive self-knowledge & Confirmed & Earlier CSM/CPPV plus C-H1: true intervention mapping outperformed a matched permuted control in 30/30 seeds. \\
Self-knowledge generalizes to unexecuted interventions & Confirmed & Fashion unmeasured repair selection and A-H4 under a strict held-out execution firewall. \\
Hidden functional organization can be recovered & Partially confirmed & Criticality, redundancy, and replaceability recovered; synergy confirmatory hypothesis failed. \\
More self-interventional experience improves self-knowledge & Confirmed & Both Experiment B primary trend hypotheses passed Holm correction. \\
Using self-knowledge can improve action & Confirmed with boundary & C3 reduced regret relative to C2, but C3 did not significantly outperform direct empirical memory C1. \\
Learned CPPV is universally superior to equal-budget Direct search & Not supported & Fashion success replicated; powered CIFAR-10/ResNet effect approximately zero. \\
SIL is a superior generic robustness method & Rejected & Dropout remained substantially stronger in powered external robustness comparisons. \\
SIL is a universal ``fourth branch'' of ML & Not established & Evidence supports an intervention-driven learning framework, not a universal taxonomy claim. \\
\bottomrule
\end{tabularx}
\end{table}

\subsection{Applied origin: predictive structural adaptation is real but domain-bounded}

The Kintsugi/CPPV lineage provided the empirical starting point for SIL. Persistent structural scars were mechanistically consequential: removing the scar mechanism reduced retention by approximately 0.70 percentage points, with the full scar condition winning in all 20 seeds in the frozen ablation family. More importantly for the later SIL formulation, intervention consequences and candidate-conditioned prospective protection were learnable rather than merely descriptive. In the independent Fashion-MNIST/ElasticCNN CPPV replication ($N=35$), Learned CPPV exceeded equal-budget Direct CPPV by 0.547 percentage points in mean damaged accuracy (31/35 wins, paired $d_z\approx0.924$, Holm-adjusted $p\approx10^{-5}$) and by 0.503 percentage points in retention (27/35 wins, $d_z\approx0.812$, Holm-adjusted $p\approx1.5\times10^{-4}$). Mean Protection Spearman was approximately 0.852, and 75.4\% of selected permanent scars had not been directly micro-trialed. Thus, the learned model extrapolated candidate value beyond the interventions used to train it.

\begin{figure}[!htbp]
\centering
\includegraphics[width=0.98\linewidth]{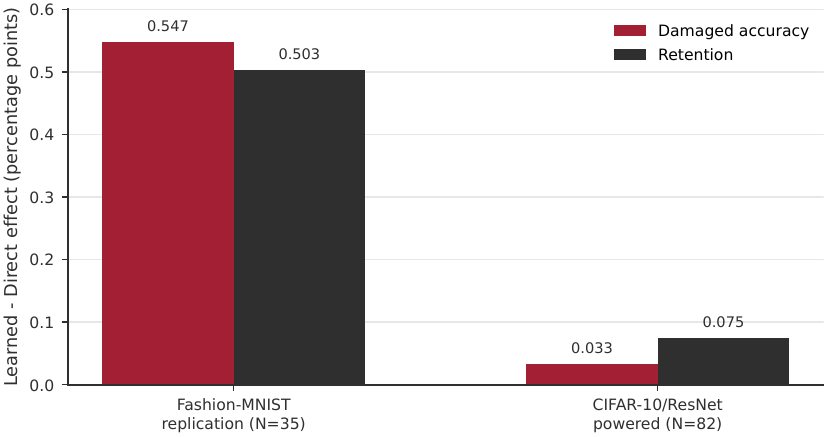}
\vspace{1mm}
\begin{minipage}{0.94\linewidth}\small
\textbf{Powered CIFAR mechanism transfer:} Protection Spearman $\approx0.688$ (81/82 positive); 78.6\% of selected scars were not directly micro-trialed; removing CSM features reduced damaged accuracy by 0.747 percentage points and retention by 0.856 percentage points relative to full Learned CPPV.
\end{minipage}
\caption{Applied CPPV evidence and the external boundary. Learned CPPV independently outperformed equal-budget Direct CPPV on Fashion-MNIST/ElasticCNN, but the corresponding powered CIFAR-10/ResNet effect was approximately zero. Predictive protection knowledge and extrapolation nevertheless transferred, separating mechanism transfer from performance superiority.}
\label{fig:applied_boundary}
\end{figure}

The powered CIFAR-10/ResNet replication established the corresponding external boundary rather than reproducing the Fashion performance advantage (Fig.~\ref{fig:applied_boundary}). Across 82 fresh seeds, Learned-minus-Direct damaged accuracy was only $+0.033$ percentage points (95\% bootstrap CI $[-0.314,+0.371]$, $d_z=0.021$, Holm $p=1.0$), while the retention difference was $+0.075$ percentage points (95\% CI $[-0.314,+0.455]$, $d_z=0.043$, Holm $p=1.0$). These near-zero effects rule out interpreting SIL/CPPV as a generally superior robustness search procedure in this architecture and domain. Dropout was also substantially stronger as a generic robustness baseline, exceeding Learned CPPV by 4.386 percentage points in damaged accuracy and 4.716 percentage points in retention.

The negative performance result did not imply failure of the predictive mechanism. Mean Protection Spearman remained approximately 0.688 and was positive in 81/82 seeds; 78.6\% of selected scars had not been directly micro-trialed. Architecture-safe consolidation preserved clean function, and removing CSM features reduced the performance of model-based repair selection: the full CSM-feature pathway improved damaged accuracy by 0.747 percentage points and retention by 0.856 percentage points relative to the no-CSM-feature ablation. The applied lineage therefore separated two claims that remained distinct throughout the remainder of the study: predictive self-interventional knowledge can transfer even when an advantage over equal-budget direct search does not.

\subsection{Experiment A: self-intervention recovers relational structure, but not completely}

Experiment A asked whether a learner could discover a construction-known functional organization that was hidden from it. The simplest part of the problem---critical-unit ranking---was recovered perfectly by pairwise SIL, with mean AUROC $=1.000$ and A-H1 passing after multiplicity correction (Table~\ref{tab:primary}). This result was not unique to SIL: direct singleton ablation, weight magnitude, and gradient/Taylor importance also reached mean AUROC $=1.000$. Criticality therefore served as a sanity check for recoverability rather than evidence that relational self-modeling was necessary.

\begin{figure}[!htbp]
\centering
\begin{minipage}[t]{0.49\textwidth}\centering
\includegraphics[width=\linewidth]{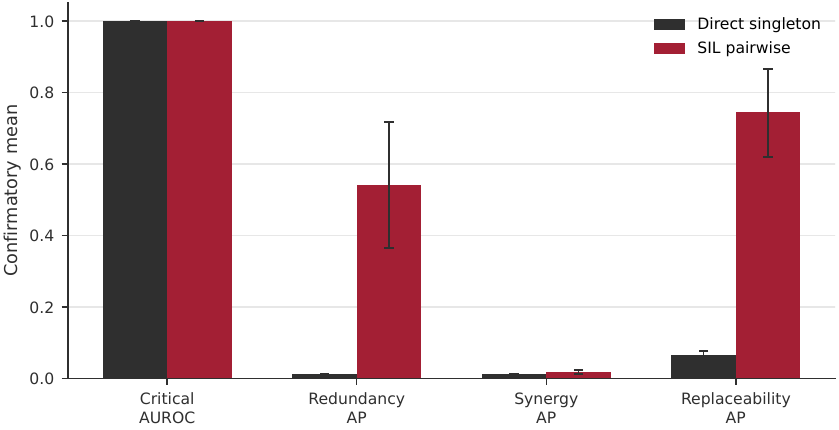}\\[-1mm]
\small (a) Ground-truth structural recovery.
\end{minipage}\hfill
\begin{minipage}[t]{0.49\textwidth}\centering
\includegraphics[width=\linewidth]{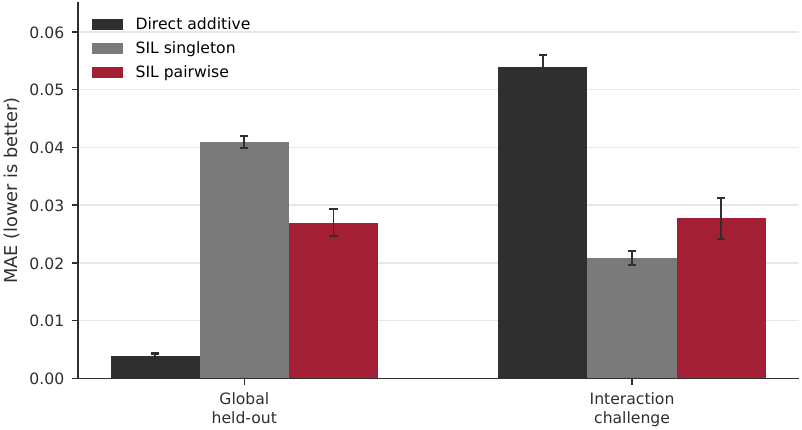}\\[-1mm]
\small (b) Prediction of genuinely unexecuted interventions.
\end{minipage}
\caption{Experiment A. Pairwise SIL strongly recovered redundancy and replaceability while synergy remained weak. Direct singleton-additive prediction was extremely accurate on the global held-out set, which was dominated by near-additive pairs, but its error increased sharply on forced interaction-sensitive challenge relations. Error bars are bootstrap 95\% confidence intervals over 20 confirmatory seeds.}
\label{fig:expA}
\end{figure}

The distinction emerged for interaction-sensitive structure (Fig.~\ref{fig:expA}a). Pairwise SIL achieved mean redundancy AP $=0.5411$, compared with $0.01087$ for the direct singleton-additive baseline. The preregistered difference was $+0.53020$ (95\% CI $[+0.35401,+0.70735]$, paired $d_z=1.293$, Holm $p=9.55\times10^{-5}$), supporting A-H2. The sign of the redundancy interaction was recovered in 58/60 true redundant relations (96.7\%). Replaceability, a secondary structural endpoint, was also recovered strongly: mean AP was $0.7463$ for pairwise SIL versus $0.0641$ for Direct, with pairwise SIL higher in all 20 seeds.

Synergy/complementarity formed the principal structural failure. Mean synergy AP was only $0.01645$ for pairwise SIL versus $0.01087$ for Direct. Although the mean preregistered difference was slightly positive ($+0.00558$), A-H3 failed after Holm correction ($p=0.1942$), precision/recall at the true-$k$ operating point were zero, and the sign of true synergy interactions was recovered in only 18.3\% of cases. Consequently, the Experiment A result is not complete functional-structure recovery; it is strong recovery of some relational motifs, especially redundancy and replaceability, together with a reproducible limitation for synergy.

Prospective prediction provided the second major result of Experiment A. Pairwise SIL reduced held-out MAE from $0.04099$ for the singleton learned model to $0.02684$, a 34.5\% reduction, with the pairwise model better in 19/20 seeds. The preregistered paired effect was $-0.01415$ (95\% CI $[-0.01633,-0.01166]$, $d_z=-2.574$, Holm $p=7.63\times10^{-6}$), confirming A-H4. Held-out Spearman increased from $0.578$ to $0.739$ and expected calibration error decreased from $0.0288$ to $0.0167$.

The direct singleton-additive predictor nevertheless had the lowest \emph{global} held-out MAE ($0.00394$), because most held-out pairs were near additive (Fig.~\ref{fig:expA}b). This advantage reversed on the forced interaction-sensitive challenge subset: Direct challenge MAE increased to $0.05384$, compared with $0.02780$ for pairwise SIL and $0.02082$ for singleton SIL. The comparison is therefore not evidence that a learned relational model universally dominates direct arithmetic composition. Instead, it shows why global prediction and structural self-knowledge must be distinguished: a simple additive rule can predict a distribution dominated by additive pairs while failing precisely on relations whose functional meaning depends on interaction.

\subsection{Experiment B: predictive self-knowledge exhibits a confirmatory learning curve}

Experiment B tested the defining learning-process prediction: if self-interventional consequences are genuine experience, increasing that experience should systematically improve predictive self-knowledge. The singleton evidence, learner, held-out set, and primary random-nested acquisition rule were held fixed; only the number of observed pairwise interventions increased through $B\in\{4,8,14,28,56\}$.

\begin{figure}[!htbp]
\centering
\begin{minipage}[t]{0.32\textwidth}\centering
\includegraphics[width=\linewidth]{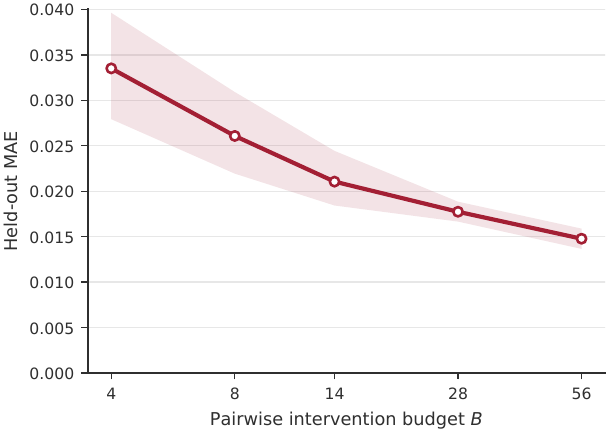}\\[-1mm]\small (a) Held-out MAE.
\end{minipage}\hfill
\begin{minipage}[t]{0.32\textwidth}\centering
\includegraphics[width=\linewidth]{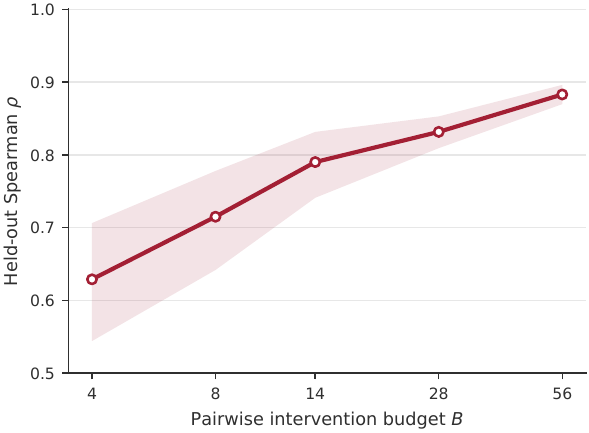}\\[-1mm]\small (b) Held-out rank correlation.
\end{minipage}\hfill
\begin{minipage}[t]{0.32\textwidth}\centering
\includegraphics[width=\linewidth]{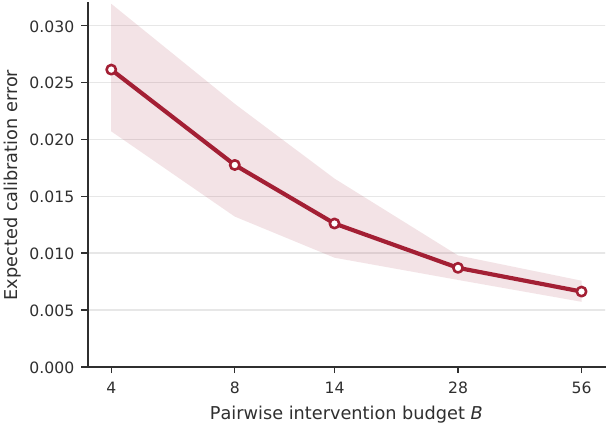}\\[-1mm]\small (c) Calibration error.
\end{minipage}
\caption{Experiment B learning curves for the preregistered random-nested acquisition policy. Singleton evidence and the held-out evaluation set were fixed while pairwise self-interventional experience increased from $B=4$ to $B=56$. Prediction error decreased, ranking quality increased, and calibration improved. Error bars are bootstrap 95\% confidence intervals over 30 fresh confirmatory seeds.}
\label{fig:expB}
\end{figure}

Both preregistered trend hypotheses passed with large margins (Table~\ref{tab:primary}; Fig.~\ref{fig:expB}). Mean held-out MAE decreased from $0.03351$ at $B=4$ to $0.01478$ at $B=56$, a 55.9\% reduction. The mean seed-level slope against $\log_2(B)$ was $-0.004841$ (95\% CI $[-0.006571,-0.003265]$, Holm $p=1.86\times10^{-8}$), with the expected negative slope in 28/30 seeds. The direct endpoint contrast $B=56-B=4$ was $-0.01873$ (95\% CI $[-0.02527,-0.01261]$, paired $d_z=-1.057$; 29/30 seeds improved).

Ranking quality improved in parallel. Mean held-out Spearman increased from $0.629$ at $B=4$ to $0.883$ at $B=56$. The preregistered Fisher-$z$ slope was $+0.15107$ (95\% CI $[+0.11304,+0.19068]$, Holm $p=2.33\times10^{-8}$), with the expected positive direction in 29/30 seeds. The raw Spearman gain from $B=4$ to $B=56$ averaged $+0.2541$ (95\% CI $[+0.1777,+0.3356]$, $d_z=1.101$; 29/30 wins).

Calibration improved as a prespecified secondary result. Expected calibration error fell from $0.02614$ to $0.00662$ across the same budget range; the ECE slope was negative in 29/30 seeds and passed the secondary Holm family ($p=5.59\times10^{-9}$). The average monotonicity score was 0.767 for MAE and 0.775 for Spearman, indicating a systematic but not artificially perfect staircase. Twenty of 30 seeds reached at least half of their final MAE improvement by $B\le14$, while 15/30 required the full $B=56$ budget to reach at least 90\% of their available best-so-far improvement.

Two secondary results constrain the interpretation. First, challenge-specific MAE decreased descriptively but its prespecified slope did not survive multiplicity correction (Holm $p=0.0767$). Second, the heuristic informed acquisition rule did not improve sample efficiency: its MAE area under the log-budget curve was on average $0.00709$ \emph{higher} than random acquisition (secondary H3, Holm $p\approx1$). The confirmatory result is therefore not that SIL already knows how to choose optimal experiments on itself. It is the more fundamental finding that increasing self-interventional experience produces systematically better predictive self-knowledge.

\subsection{Experiment C: correct self-knowledge improves action when used, but not beyond every direct policy}

\begin{figure}[!htbp]
\centering
\begin{minipage}[t]{0.32\textwidth}\centering
\includegraphics[width=\linewidth]{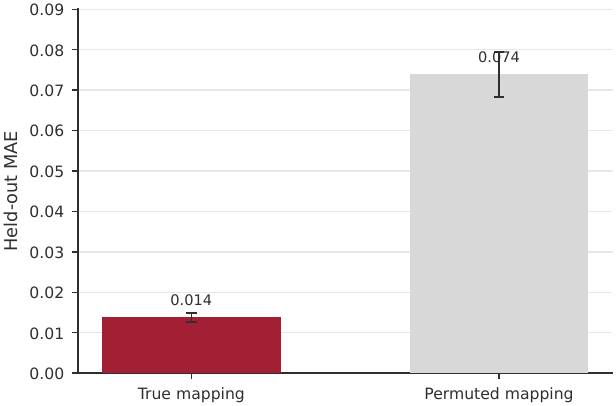}\\[-1mm]\small (a) True vs. permuted self-model.
\end{minipage}\hfill
\begin{minipage}[t]{0.32\textwidth}\centering
\includegraphics[width=\linewidth]{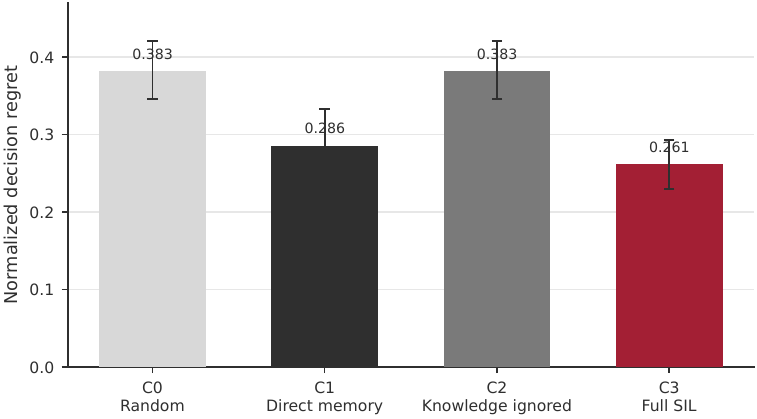}\\[-1mm]\small (b) Causal action arms.
\end{minipage}\hfill
\begin{minipage}[t]{0.32\textwidth}\centering
\includegraphics[width=\linewidth]{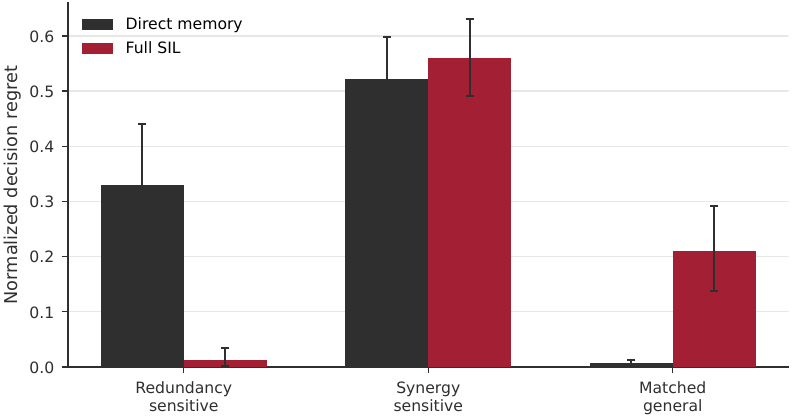}\\[-1mm]\small (c) Direct memory vs. Full SIL by episode family.
\end{minipage}
\caption{Experiment C. Preserving the intervention--consequence mapping substantially improved prospective prediction under the matched control. Coupling the same learned model to action (C3) reduced regret relative to ignoring it (C2), while Full SIL did not significantly outperform the direct empirical-memory policy (C1). The decomposition shows why: relational self-knowledge was particularly effective for redundancy-sensitive decisions, whereas direct singleton-additive memory remained stronger on matched-general cases and synergy remained difficult. Error bars are bootstrap 95\% confidence intervals over 30 confirmatory seeds.}
\label{fig:expC}
\end{figure}

Experiment C separated the causal links from intervention evidence to predictive knowledge and from predictive knowledge to action. The first contrast destroyed the semantics of the pairwise experience while preserving its quantity and marginal target distribution. The true self-model achieved mean held-out MAE $=0.01383$, whereas the matched pair-target-permuted model reached $0.07396$ (Fig.~\ref{fig:expC}a). The paired difference was $-0.06013$ (95\% CI $[-0.06524,-0.05502]$, $d_z=-4.105$), with 30/30 seeds favoring the true mapping and Holm $p=2.79\times10^{-9}$. Mean held-out Spearman was $0.837$ for the true model versus $0.114$ for the permuted control. C-H1 therefore provides direct negative-control evidence that prospective prediction depends on the intervention--consequence mapping rather than merely on exposure to the target distribution.

The second causal contrast used the same fitted self-model and the same candidate sets in two conditions. C2 retained the model but ignored its predictions and used the precommitted random action; C3 used the model predictions to select the action. C0 and C2 were verified to make identical random choices episode-by-episode. Mean normalized regret was consequently identical for C0 and C2 ($0.38256$), whereas Full SIL C3 reduced regret to $0.26140$ (Fig.~\ref{fig:expC}b). The preregistered C3-minus-C2 effect was $-0.12116$ (95\% CI $[-0.17212,-0.07035]$, $d_z=-0.844$, 24/30 wins, Holm $p=2.32\times10^{-4}$). This matched ablation isolates the downstream role of the self-model: the same learned knowledge improved decisions when it controlled action.

The stronger claim that learned model-guided action should outperform direct empirical memory was not supported. C1, which used singleton-additive intervention memory without a learned relational model, achieved mean regret $0.28588$, close to C3's $0.26140$. The preregistered C3-minus-C1 effect was $-0.02448$ (95\% CI $[-0.07539,+0.02980]$; 19 wins, 11 losses; $d_z=-0.162$; Holm $p=0.2449$), so C-H3 failed. C1 also selected the exact-best action more often (33.0\% vs. 27.0\%) and achieved a higher top-2 rate (62.6\% vs. 47.8\%), despite C3's lower mean regret.

The episode-family decomposition explains this boundary (Fig.~\ref{fig:expC}c). On redundancy-sensitive decisions, Full SIL was almost oracle-optimal (mean regret $0.0124$) while direct empirical memory remained substantially worse ($0.3292$). On matched-general decisions, the relation reversed: the direct singleton-additive rule was nearly perfect ($0.0064$) while C3 reached $0.2110$. Synergy-sensitive decisions remained difficult for both policies, with regret $0.5220$ for C1 and $0.5609$ for C3. The action result therefore recapitulates the structural findings from Experiment A: relational self-knowledge is particularly useful when the decision depends on non-additive redundancy, but it is not a universal replacement for simpler direct empirical rules and does not resolve the observed weakness for synergy.

\subsection{Integrated evidence}

Across the three final mechanistic experiments, seven of nine preregistered primary hypotheses passed after their frozen Holm corrections. Counting primary hypotheses is not itself the inferential argument; the important result is the pattern across independent tests. Experiment A showed that self-interventions can reveal nontrivial relational functional structure and support prospective prediction, while also exposing a failure to recover synergy. Experiment B provided confirmatory evidence that predictive self-knowledge improves systematically with additional self-interventional experience. Experiment C showed that preserving the correct intervention--consequence mapping materially improves prospective prediction and that the same learned self-model can reduce decision regret when its predictions are actually used. At the same time, C3 did not outperform direct empirical memory, and the powered CIFAR-10/ResNet study did not reproduce the Fashion-MNIST Learned-vs-Direct robustness advantage.

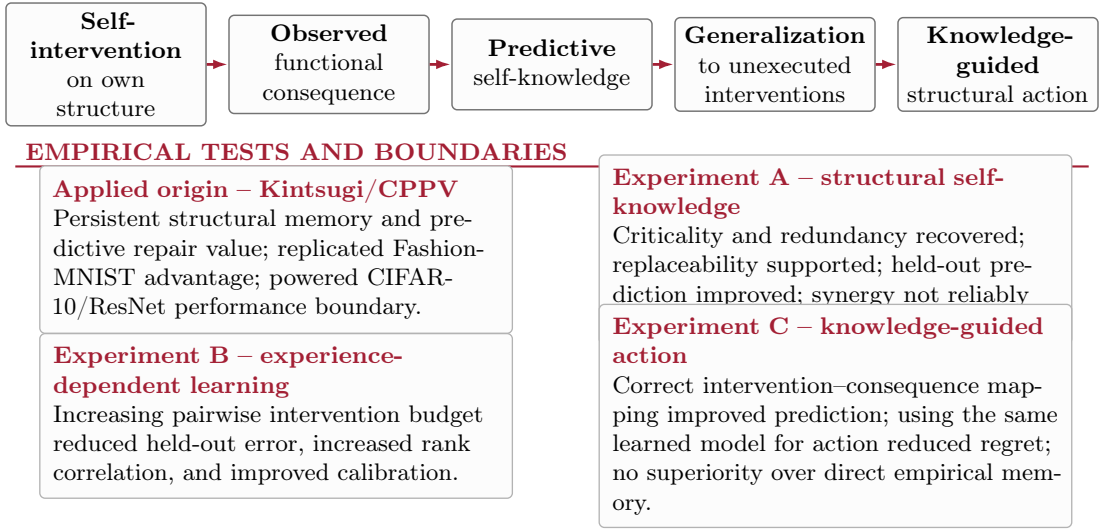
\begin{figure}[!htbp]
\centering
\definecolor{MITRed}{HTML}{A31F34}
\begin{tikzpicture}[
    font=\footnotesize,
    process/.style={draw=black!55, fill=black!1, rounded corners=2pt, align=center,
                   text width=24mm, minimum height=12mm, inner sep=3.5pt, line width=.55pt},
    evidence/.style={draw=black!30, fill=black!1, rounded corners=2pt, align=left,
                    text width=59mm, minimum height=17mm, inner sep=5pt, line width=.5pt},
    arr/.style={-{Latex[length=2mm,width=1.3mm]}, line width=1.05pt, draw=MITRed}
]
\node[process] (i) at (0.8,0) {\textbf{Self-intervention}\\on own structure};
\node[process] (c) at (3.75,0) {\textbf{Observed}\\functional consequence};
\node[process] (k) at (6.70,0) {\textbf{Predictive}\\self-knowledge};
\node[process] (g) at (9.65,0) {\textbf{Generalization}\\to unexecuted interventions};
\node[process] (a) at (12.60,0) {\textbf{Knowledge-guided}\\structural action};

\draw[arr] (i.east) -- (c.west);
\draw[arr] (c.east) -- (k.west);
\draw[arr] (k.east) -- (g.west);
\draw[arr] (g.east) -- (a.west);

\node[anchor=west, text=MITRed, font=\bfseries\footnotesize] at (-0.4,-1.15) {EMPIRICAL TESTS AND BOUNDARIES};
\draw[MITRed, line width=.8pt] (-0.4,-1.35) -- (13.8,-1.35);

\node[evidence] (origin) at (3.05,-2.45) {\textcolor{MITRed}{\textbf{Applied origin -- Kintsugi/CPPV}}\\
Persistent structural memory and predictive repair value; replicated Fashion-MNIST advantage; powered CIFAR-10/ResNet performance boundary.};

\node[evidence] (ea) at (10.45,-2.45) {\textcolor{MITRed}{\textbf{Experiment A -- structural self-knowledge}}\\
Criticality and redundancy recovered; replaceability supported; held-out prediction improved; synergy not reliably recovered.};

\node[evidence] (eb) at (3.05,-4.65) {\textcolor{MITRed}{\textbf{Experiment B -- experience-dependent learning}}\\
Increasing pairwise intervention budget reduced held-out error, increased rank correlation, and improved calibration.};

\node[evidence] (ec) at (10.45,-4.65) {\textcolor{MITRed}{\textbf{Experiment C -- knowledge-guided action}}\\
Correct intervention--consequence mapping improved prediction; using the same learned model for action reduced regret; no superiority over direct empirical memory.};
\end{tikzpicture}
\caption{Self-Interventional Learning (SIL) and the empirical evidence map. The upper pathway defines the operational mechanism tested in this work. The lower evidence grid summarizes the applied Kintsugi/CPPV origin and the three mechanistic experiments, including their principal negative boundaries rather than depicting SIL as universally superior.}
\label{fig:sil_map}
\end{figure}

The resulting evidence therefore supports the mechanism shown in Fig.~\ref{fig:sil_map}: self-intervention can generate training experience about the learner's own functional organization; that experience can be converted into predictive self-knowledge; the quality of that knowledge can improve with additional experience; it can generalize to unexecuted interventions; and it can influence subsequent action. The evidence does not support complete structural self-understanding, universal superiority over direct intervention strategies, or a general robustness advantage.

\section{Discussion}
\label{sec:discussion}

\subsection{Can neural networks learn by experimenting on themselves?}

The title of this paper poses a deliberately operational question. The experiments do not address phenomenal self-awareness, introspection, or a general-purpose representation of the self. They ask whether controlled interventions on a learner's own functional substrate can generate experience from which the learner acquires predictive knowledge about that substrate, whether the quality of that knowledge improves with additional experience, whether it generalizes to interventions not executed during self-discovery, and whether the acquired knowledge can influence subsequent action. Under this definition, the answer supported by the present evidence is yes, with important qualifications.

Experiment A established that intervention experience can reveal functional structure that is not reducible to ordinary scalar importance. Critical components were easy for several methods to identify, including classical magnitude and gradient-based rankings, and therefore do not by themselves motivate a new learning framework. The more diagnostic results concerned relational structure. Pairwise SIL strongly recovered redundancy and replaceability, and its learned predictions generalized to genuinely unexecuted interventions. At the same time, synergy recovery failed its preregistered confirmatory hypothesis. The resulting self-model was therefore neither trivial nor complete: it captured some forms of relational organization while systematically missing another.

Experiment B supplied the strongest evidence that this mechanism should be interpreted as learning rather than as a fixed diagnostic. With singleton evidence, learner architecture, held-out set, and acquisition rule fixed, increasing only the number of observed pairwise self-interventions produced a systematic reduction in held-out prediction error, an increase in ranking quality, and improved calibration. These effects appeared across nearly all fresh confirmatory seeds and were tested through preregistered seed-level trends rather than a favorable post-hoc endpoint contrast. The resulting intervention-budget curve is conceptually important because it links predictive self-knowledge to accumulated experience: the learner did not merely possess a score assigned to its components; it became better at predicting itself as it observed more consequences of its own interventions.

Experiment C then separated possession of self-knowledge from use of self-knowledge. Permuting pairwise intervention consequences while preserving the amount of evidence and its marginal target distribution sharply degraded held-out prediction, demonstrating that the useful signal was carried by the mapping between intervention identity and functional consequence. More importantly, the C2/C3 contrast held the intervention history, fitted model, predictions, and candidate set fixed and changed only whether those predictions controlled the action. Using the self-model reduced decision regret relative to ignoring the same self-model. This provides causal evidence for a downstream role of the learned predictions under the tested decision protocol.

The combined evidence therefore supports an operational form of self-interventional learning:
\begin{center}
\resizebox{0.96\linewidth}{!}{%
$\text{self-intervention}\rightarrow\text{observed consequence}\rightarrow\text{predictive self-knowledge}\rightarrow\text{experience-dependent improvement}\rightarrow\text{held-out generalization}\rightarrow\text{knowledge-guided action}$}
\end{center}
The claim is deliberately narrower than self-awareness and broader than post-hoc feature attribution. What is learned is a predictive model of functional consequences under interventions on the learner's own substrate.

\subsection{Self-modeling, mechanistic interpretability, and the locus of the experimenter}

The closest conceptual precedents come from several research traditions that place different entities in the role of experimenter. In embodied self-modeling, robots learn predictive descriptions of their own morphology or dynamics and use those models to adapt, plan, or recover from damage \cite{bongard2006resilient,kwiatkowski2019taskagnostic,chen2022visual,ledezma2023selfdiscovery}. This line established that a machine can benefit from a learned model of itself rather than only of its environment. SIL shares the predictive and action-oriented character of that work, but moves the object of self-modeling from embodied morphology or dynamics to the functional organization of a neural computation and makes controlled structural intervention the source of training evidence.

Neural self-modeling provides a second point of contact. Premakumar et al. showed that training networks to predict their own internal states as an auxiliary task can alter the networks themselves, reducing several measures of complexity and increasing regularity \cite{premakumar2024unexpected}. That result demonstrates that self-prediction can become part of the learning objective of a neural system. SIL differs in the target of prediction: rather than predicting the current internal state, it predicts the functional consequence of an intervention on internal structure. This distinction introduces a counterfactual dimension---what would happen if a component or combination of components were altered---and makes held-out intervention prediction the central generalization test.

Mechanistic interpretability occupies a complementary position. Ablation, activation patching, interchange interventions, causal tracing, and circuit-discovery methods use interventions to infer how neural components contribute to behavior \cite{geiger2023causal,geiger2022iit,conmy2023automated,heimersheim2024patching}. In these methods, however, the intervention evidence is normally consumed by an external analyst or interpretability algorithm. SIL changes the locus of the learner: the intervention evidence becomes experience for the system's own predictive self-model. This is not a claim that SIL replaces mechanistic interpretability. The two can instead be viewed as different uses of a common experimental instrument. Mechanistic interpretability asks what an external investigator can infer from interventions; SIL asks what a learner can acquire and use when intervention consequences become part of its own experience.

This distinction also clarifies why the synthetic ground-truth experiment was important for separating scalar importance from relational self-knowledge. A method can rank important units without learning relational structure. Recent analysis of activation patching shows that component-wise mediation scores can contain or miss interaction effects and that pairwise or higher-order interactions may be required when causal importance depends on the state of other components \cite{vaidyanathan2026interactions}. The redundancy result in Experiment A is consistent with this broader methodological point: a pair can be functionally consequential even when its members look individually unimportant. The present study goes beyond diagnosing that interaction by asking whether the learner can use intervention experience to predict it. At the same time, the failure on synergy is a reminder that representing interaction-sensitive functional organization remains substantially harder than detecting isolated importance.

\subsection{Relational self-knowledge is useful where additivity fails}

One of the most informative findings is that learned relational self-knowledge was not uniformly better than direct empirical memory. In Experiment A, the direct singleton-additive predictor achieved the lowest global held-out MAE because most held-out pairs were nearly additive. Its advantage disappeared on the forced interaction-sensitive challenge subset. In Experiment C, the same pattern reappeared at the decision level. Full SIL was close to oracle-optimal for redundancy-sensitive episodes, while the direct singleton-additive policy was nearly perfect on matched-general episodes. The preregistered Full-SIL-versus-direct-memory contrast consequently did not reach significance.

This is not a contradiction between Experiments A and C; it is the central boundary they jointly reveal. A learned relational self-model is most valuable when the decision depends on structure that cannot be reconstructed by adding singleton effects. If the relevant intervention consequences are approximately additive, a simple empirical rule can be both more accurate and more data-efficient. Complexity is therefore not intrinsically beneficial. The appropriate interpretation is conditional: relational self-knowledge adds value when relational structure matters.

The redundancy results provide the clearest example. A redundant pair has the characteristic that individual lesions can be mild while a joint lesion is severe. No scalar importance ranking can express this relation without additional interaction information. Pairwise SIL recovered the sign of 58/60 true redundant interactions in Experiment A, and Full SIL achieved near-zero regret on redundancy-sensitive decisions in Experiment C. This convergence across structural recovery, prospective prediction, and action is stronger evidence than any one metric alone.

Synergy provides the corresponding negative example. Experiment A failed to recover synergistic structure reliably, and synergy-sensitive decisions remained difficult in Experiment C. The same learner that successfully modeled redundancy therefore did not simply acquire a generic interaction detector. The asymmetry suggests that the representation, intervention distribution, or function approximator captured some interaction geometries better than others. Because the protocols were frozen before confirmatory evaluation, the present study does not identify which explanation is correct. Resolving that question belongs to future work rather than to a post-hoc rescue of the current evidence.

\subsection{Experience quantity and the limits of intervention acquisition}

The learning curve in Experiment B has two implications. First, it provides evidence for an experience-dependent self-model: predictive quality changes systematically as intervention experience accumulates. Second, it separates the existence of a learning curve from the problem of choosing informative interventions. The primary random-nested acquisition policy produced a strong learning curve, whereas the prespecified heuristic informed policy did not improve sample efficiency and was often worse at small budgets.

This distinction mirrors a central problem in active causal discovery, where selecting where and how to intervene is itself an experimental-design problem \cite{tigas2022interventions}. In SIL, the same issue arises internally: once a learner is permitted to experiment on itself, it must eventually decide which interventions are worth performing. The negative informed-acquisition result shows that this problem is not solved by a simple heuristic combining similarity, activation correlation, singleton consequence, and random exploration. A heuristic that prioritizes apparently informative pairs can distort the training distribution and reduce global predictive efficiency. The current evidence therefore supports self-interventional learning under random exploration, not optimal self-experiment design.

This boundary is useful because it prevents two distinct claims from being conflated. Experiment B provides confirmatory evidence that more intervention experience improves self-knowledge under the tested protocol. It does not demonstrate that the learner can already choose the best experiments on itself. A mature SIL system would ideally require both capabilities: a predictive self-model and an acquisition policy that selects interventions according to expected information value, safety, cost, and decision relevance. The latter remains open.

\subsection{From Kintsugi and CPPV to a broader learning framework}

The Kintsugi/CPPV lineage provides an applied instantiation of the same logic under structural adaptation. Persistent scars stored consequences of earlier perturbations, candidate-conditioned protection values were learned from controlled micro-trials, and the learned model extrapolated to repair candidates not directly evaluated. On Fashion-MNIST/ElasticCNN, this predictive mechanism also yielded a replicated performance advantage over equal-budget direct local search. The powered CIFAR-10/ResNet replication then established an essential boundary: prediction and extrapolation transferred, but performance superiority did not.

That distinction motivated the final experiments. If SIL were defined by robustness improvement, the powered CIFAR result would weaken the framework directly. If SIL is defined instead by intervention-driven acquisition of predictive self-knowledge, the external result becomes more informative: a self-model can remain predictive and operationally useful without necessarily beating direct search on the downstream performance metric. The final A/B/C program tests precisely this broader mechanism independently of robustness.

The resulting interpretation is therefore not that Kintsugi and CPPV were replaced by a synthetic demonstration. They serve different evidential roles. Kintsugi/CPPV shows that self-interventional knowledge can arise in a practical structural-adaptation setting and can sometimes improve downstream resilience. The synthetic experiments supply the causal and ground-truth controls required to determine what kind of knowledge has actually been acquired. Together they motivate SIL as a framework rather than as a single repair algorithm.

\subsection{What the present evidence does and does not establish}

The evidence supports four claims with different strengths. First, functional consequences of controlled interventions on a neural system can be learned prospectively. This is supported across the earlier CSM/CPPV experiments, Experiment A held-out prediction, Experiment B learning curves, and the Experiment C permutation control. Second, the learned model can contain relational information that scalar importance or additive singleton memory does not capture, most clearly for redundancy. Third, predictive self-knowledge improves with additional intervention experience. Fourth, coupling learned self-knowledge to action can reduce regret relative to an otherwise identical condition that ignores it.

Several stronger claims are not supported. SIL does not recover complete functional organization; synergy remains a clear failure mode. The learned model is not universally more accurate than direct empirical composition, nor is model-guided action universally better than direct empirical action. The heuristic informed acquisition policy did not improve sample efficiency. The challenge-specific learning-curve endpoint did not survive multiplicity correction. The powered CIFAR-10/ResNet study found approximately zero Learned-versus-Direct robustness advantage, and dropout remained substantially stronger as a generic robustness baseline. These findings are part of the result, not exceptions to it.

Finally, the term \emph{self-knowledge} is used here in a strictly predictive and operational sense. It denotes learned information about the functional consequences of interventions on the system's own substrate. It does not imply consciousness, phenomenology, subjective introspection, or a privileged epistemic status. Likewise, the results do not justify describing SIL as a universal new ``branch'' of machine learning. They support a narrower proposition: self-intervention can define a learnable source of predictive information about a system's own functional organization.

\section{Limitations and Future Work}
\label{sec:limitations}

The present study deliberately trades breadth for mechanistic control. Several limitations therefore define the scope of the conclusions.

\paragraph{Synthetic structural ground truth.}
The strongest causal tests use a deliberately constructed 24-component system in which criticality, redundancy, replaceability, synergy, and peripheral roles are known by design. This construction makes structural recovery measurable and protects against circular interpretation, but it does not establish that the same role taxonomy or recovery quality will hold in modern large-scale networks. The applied Fashion-MNIST and CIFAR-10 experiments provide evidence outside the synthetic setting, yet they test repair-value prediction rather than construction-known functional roles. Extending ground-truth-style evaluation to larger and more heterogeneous architectures remains an important challenge.

\paragraph{Incomplete interaction modeling.}
Synergy was the most consistent weakness. The preregistered synergy hypothesis failed in Experiment A, synergy interaction signs were recovered poorly, and synergy-sensitive decisions remained difficult in Experiment C. The current experiments do not distinguish whether this failure is caused primarily by the feature representation, learner capacity, acquisition distribution, definition of the intervention consequence, or the geometry of the constructed synergy motif. Any method designed specifically after observing this failure would constitute a new experiment and is therefore outside the frozen evidence reported here.

\paragraph{Direct empirical strategies remain strong.}
The learned relational self-model did not universally dominate simpler alternatives. Direct singleton-additive prediction had the lowest global held-out MAE in Experiment A because the evaluation distribution contained many approximately additive pairs. In Experiment C, Full SIL did not significantly outperform the direct empirical-memory policy and had lower exact-best and top-2 action rates despite lower mean regret. SIL should therefore not be interpreted as a mandatory replacement for direct intervention memory. A practical system may need to arbitrate between direct, additive, and model-based estimates according to the interaction structure and uncertainty of the current decision.

\paragraph{Intervention acquisition is unresolved.}
Experiment B establishes that more self-interventional experience improves predictive self-knowledge, but the prespecified informed acquisition heuristic did not improve sample efficiency over random nested exploration. The work therefore demonstrates learning from interventions, not an optimal strategy for choosing interventions. Future systems would need acquisition criteria that account for information gain, coverage, intervention cost, risk, and downstream decision value. Such criteria should be evaluated prospectively rather than tuned retrospectively to the present learning curves.

\paragraph{Restricted action space and intervention vocabulary.}
The final causal experiment uses discrete two-component reconfiguration choices and a finite intervention vocabulary consisting primarily of lesions and substitutions. Real neural systems admit continuous interventions, parameter edits, routing changes, activation manipulations, module replacement, and interventions at multiple representational scales. Whether SIL can learn across such heterogeneous intervention families remains unknown. The current evidence is strongest for the intervention classes explicitly tested.

\paragraph{Architecture and domain transfer.}
The applied external validation exposes a substantive transfer boundary. On CIFAR-10/ResNet, candidate-conditioned protection remained predictable and model-based selection extrapolated to unmeasured candidates, but Learned CPPV did not outperform equal-budget Direct CPPV in the powered $N=82$ replication. The cross-domain evidence therefore supports transfer of predictive mechanism more strongly than transfer of downstream performance advantage. Additional domains may ultimately reveal where each level of transfer holds, but no further benchmark expansion was performed after the prespecified final experiments.

\paragraph{Scale and computational economy.}
The final mechanistic experiments were intentionally small enough to permit many fresh-seed replications, strict intervention firewalls, and construction audits. They do not address the computational cost of self-intervention in foundation-scale models, where executing interventions may be expensive and some structural changes may be unsafe or irreversible. Practical scaling will require surrogate evaluation, reversible interventions, uncertainty-aware stopping, and explicit intervention budgets.

\paragraph{Operational meaning of self-knowledge.}
The term \emph{self-knowledge} refers only to predictive knowledge of intervention consequences on the system's own functional substrate. It should not be conflated with consciousness, self-awareness, agency, or human-like introspection. The experiments demonstrate a functional learning property, not a phenomenological one.

\paragraph{Future work.}
The most natural extensions follow directly from the observed boundaries rather than from attempts to improve the frozen results: interaction representations that can distinguish redundancy from synergy; principled active self-experiment design; uncertainty-aware selection between direct memory and learned prediction; heterogeneous and continuous intervention spaces; multi-scale self-models for deeper architectures; and decision tasks in which the value of relational self-knowledge can be assessed under realistic intervention cost and safety constraints. These are proposed as new research directions, not as missing experiments required to support the claims of the present study.

\section{Conclusion}
\label{sec:conclusion}

This study asked whether neural networks can learn by experimenting on themselves. In the operational sense tested here, the evidence supports a qualified affirmative answer. Controlled interventions on a neural system's own functional substrate produced consequences that could be learned prospectively. The resulting self-model recovered nontrivial relational structure, generalized to interventions not executed during self-discovery, improved systematically as self-interventional experience increased, and reduced decision regret when its predictions were actually coupled to action.

The evidence also defines clear limits. Structural self-knowledge was incomplete: redundancy and replaceability were recovered far more reliably than synergy. More self-interventional experience improved prediction, but a simple informed acquisition heuristic did not improve sample efficiency. Model-guided action outperformed an otherwise identical condition that ignored the same self-model, yet it did not significantly outperform direct empirical memory. In the applied Kintsugi/CPPV lineage, predictive protection knowledge transferred from Fashion-MNIST/ElasticCNN to CIFAR-10/ResNet, whereas the performance advantage over equal-budget direct repair search did not.

These results motivate \emph{Self-Interventional Learning} as a framework in which intervention on the learner becomes a source of learning experience about the learner. Its defining object is not merely the current internal state, and its goal is not merely post-hoc explanation. SIL learns a predictive mapping from interventions on internal functional structure to their consequences and can use that mapping prospectively. The present evidence does not establish complete self-understanding or universal algorithmic superiority. It supports a more specific principle: a neural system can acquire useful predictive knowledge about its own functional organization by experimentally perturbing itself, and that knowledge can improve with experience and guide later action.

\section*{Data and Code Availability}
The frozen protocols, runnable implementations, and lightweight confirmatory result artifacts for Experiments A--C are included with the reproducibility appendix accompanying this submission. The applied studies use Fashion-MNIST and CIFAR-10, which are publicly available benchmark data sets. Large model checkpoints are not required for the synthetic A--C analyses and are intentionally excluded from the lightweight archive. The submission package also contains the frozen evidence audit for the powered CIFAR-10/ResNet replication.

\vskip 0.2in
\setlength{\bibsep}{0.5pt}
\bibliography{references_master_y8}
\end{document}